\documentclass[10pt,twocolumn,letterpaper,pagenumbers]{article}

\usepackage{wacv}              
\usepackage{multirow}
\usepackage{graphicx}
\usepackage{amsmath}
\usepackage{amssymb}
\usepackage{booktabs}
\allowdisplaybreaks

\usepackage[pagebackref,breaklinks,colorlinks]{hyperref}

\usepackage[capitalize]{cleveref}
\crefname{section}{Sec.}{Secs.}
\Crefname{section}{Section}{Sections}
\Crefname{table}{Table}{Tables}
\crefname{table}{Tab.}{Tabs.}

\def\wacvPaperID{604} 
\def\confName{WACV}
\def\confYear{2025}

\begin{document}

\title{Solar Multimodal Transformer: Intraday Solar Irradiance Predictor using Public Cameras and Time Series\thanks{Accepted to WACV2025}}

\author{Yanan Niu\textsuperscript{1}, Roy Sarkis\textsuperscript{1}, Demetri Psaltis\textsuperscript{1}, Mario Paolone\textsuperscript{1}, Christophe Moser\textsuperscript{1}, Luisa Lambertini\textsuperscript{1, 2}\\
\textsuperscript{1}EPFL, Lausanne, 1015, Vaud, Switzerland\\
\textsuperscript{2}Universita’ della Svizzera Italiana (USI), Lugano, 6900, Ticino, Switzerland\\
{\tt\small yanan.niu@epfl.ch}
}

\maketitle

\begin{abstract}
   Accurate intraday solar irradiance forecasting is crucial for optimizing dispatch planning and electricity trading. For this purpose, we introduce a novel and effective approach that includes three distinguishing components from the literature: 1) the uncommon use of single-frame public camera imagery; 2) solar irradiance time series scaled with a proposed normalization step, which boosts performance; and 3) a lightweight multimodal model, called Solar Multimodal Transformer (SMT), that delivers accurate short-term solar irradiance forecasting by combining images and scaled time series. Benchmarking against Solcast, a leading solar forecasting service provider, our model improved prediction accuracy by 25.95\%. Our approach allows for easy adaptation to various camera specifications, offering broad applicability for real-world solar forecasting challenges.
\end{abstract}

\section{Introduction}
\label{sec:intro}
The inherent uncertainty in solar energy supply poses significant challenges in managing sustainable systems. While solar irradiance forecasting is well-developed for long-term, day-ahead, and very short-term intervals up to 10 minutes, the intermediate range—from 10 minutes to a few hours—remains under-studied. This prediction horizon is crucial for electricity markets~\cite{paletta2023advances}. For instance, entities with access to the intraday market can use these forecasts for strategic trading to manage operating costs effectively. These forecasts allow for informed decisions on trading positions, which can be adjusted from 15 minutes to several hours before energy delivery, depending on country-specific regulations and market conditions. Beyond these applications, the impact of accurate solar forecasting can extend to advanced building design~\cite{Dong2021}, urban planning~\cite{Tascikaraoglu2018}, climate studies~\cite{Guermoui2018}, and economic analyses~\cite{de2015error,zhang2015baseline}, among others.

Such forecasting can incorporate a variety of data modalities. This includes endogenous data, such as time series of historical global horizontal irradiance (GHI) measurements, as well as exogenous data like meteorological variables (air temperature, wind speed, humidity, etc.), all-sky images, and satellite imagery. Historically, time series models using GHI have dominated the field due to their simplicity. These models, however, rely solely on historical data and do not incorporate real-time exogenous information. In contrast, advanced deep learning techniques has popularized image-based methods, which utilize sky images to provide direct visual cues for near-future weather conditions. Among them, all-sky images and their corresponding time-lapse videos are extensively utilized in solar forecasting. These resources are more readily accessible than satellite imagery and offer practical advantages due to their manageable sizes and familiar, universally applicable formats, which simplify the deployment of predictive models for localized forecasting. Convolutional neural networks (CNNs) serve as feature extractors for frames, while recurrent neural networks (RNNs), like long short-term memory networks (LSTMs), create temporal links in video-based forecasting~\cite{Talha2019}. The development of transformer-based models, such as Vision Transformers (ViTs), has demonstrated their robust capabilities beyond natural language processing (NLP), extending into computer vision (CV) and multimodal interactions. Gao \etal~\cite{Gao2022} have successfully utilized ViTs to process sequences of all-sky images, thereby enhancing the accuracy and efficiency of solar forecasting.

The use of professional all-sky cameras or satellite images, with their multi-frame settings, presents challenges for scalability. To provide a solution that can be widely implemented while maintaining high accuracy, we propose the use of exogenous single-frame public camera images and endogenous GHI time series for short-term forecasting over several hours. To merge different data modalities effectively, we introduce a lightweight transformer-based model: the Solar Multimodal Transformer (SMT).

Compared to all-sky cameras or satellite images, public cameras are more accessible, thereby enhancing scalability. Any camera that captures a broader view, including both sky and ground scenes, can be utilized for forecasting. In fact, public cameras have demonstrated superior performance in our experiments. Unlike all-sky cameras dedicated solely to sky observations, the broader view from public cameras provides additional data, such as reflections and shadows, which are essential for a comprehensive analysis, as shown in \cref{sec:patch}. Furthermore, the use of fish-eye lenses in all-sky cameras, combined with their horizontal mounting, makes them particularly susceptible to image distortion from accumulations of dew, dust, or other contaminants, leading to less informative inputs for forecasting.

Existing models that rely on time-lapse videos or multiple frames often face challenges, such as increased technical demands on camera technology, or issues where trained models become inoperative when missing frames are present. To streamline the model’s architecture and overcome these challenges, our SMT utilizes single-frame inputs and incorporates historical GHI time series to replace the temporal information typically embedded in sequential frames. The transformer component in SMT facilitates an early and deep integration of spatial and temporal data from various modalities, enhancing forecasting accuracy by making full use of this straightforward temporal information.

In order to boost the performance of SMT, we introduce a normalization step that facilitates the integration of time series with other data modalities. Specifically, we compute the theoretical GHI using clear sky models as no-cloud reference~\cite{ineichen2006comparison, Haurwitz1945}. Each actual GHI measurement is then scaled by its corresponding maximal clear sky value for that day. This process detrends the data throughout the year, shifting the model's focus to predicting the ``degree of clear sky" rather than absolute GHI values.

Overall, our approach significantly outperforms several baseline single-modality approaches~\cite{Talha2019, Gao2022,Lai2017,Sarkis2024}, enhancing computational efficiency, scalability, and maintaining high accuracy, with a 25.95\% increase in forecast accuracy over industry leader Solcast. Our codes are publicly available\footnote{\url{https://github.com/YananNiu/SMT}}. Our contributions can be summarized as follows:
\begin{itemize}
\item Data selection strategy: We demonstrate the simplicity and effectiveness of leveraging single-frame images from public cameras for solar forecasting, using GHI time series to provide historical temporal information.  
\item Preprocessing step: We introduce a normalization step for GHI time series that significantly enhances the forecasting performance.
\item Model design: We propose the SMT, which effectively merges data from diverse modalities through early fusion, elevating the predictive accuracy to a new level.
\item Model interpretation: We reveal how different data modalities vary in contribution to forecasting as the weather changes through attention analysis of SMT.
\end{itemize}
\section{Related work}

Traditionally, short-term (hours) solar forecasting has been conducted using analytical equations, such as the numerical weather prediction (NWP)~\cite{Bi2023}. Here, we concentrate on indirect statistical and machine learning methods.

\textbf{Time series models.} For solar forecasting utilizing endogenous historical GHI data, a diverse range of statistical and deep learning models are available for selection. Statistical methods, such as exponential smoothing (ETS) and autoregressive integrated moving average (ARIMA) models—along with their extension, seasonal auto-regressive integrated moving average (SARIMA)—are well-suited for non-stationary time series forecasts. Sarkis \etal~\cite{Sarkis2024} demonstrated that ARIMA models were less effective compared to their image-based approach for predicting GHI. Additionally, RNNs and their variants such as LSTM~\cite{hochreiter1997long} and gated recurrent units (GRU)~\cite{chung2014empirical} have surpassed traditional statistical models by capitalizing on cross-series data and larger datasets~\cite{Hewamalage2021}. CNNs also prove advantageous in handling time series data due to their ability to manage shift-invariant features across various time scales~\cite{lecun1995convolutional}. Lai \etal~\cite{Lai2017} developed the Long- and Short-Term Time-series Network (LSTNet) to forecast GHI across multiple photovoltaic installations, incorporating CNN modules to extract short-term patterns and GRU layers for feature aggregation. Additionally, a recurrent-skip component was integrated to address very long-term patterns, enhancing the efficiency and stability of the model's training convergence. 

\textbf{Image-based models.} In the realm of solar forecasting using image data, tasks are typically framed as regression problems. This approach is distinct from, yet complements, conventional 2D CV tasks such as image classification, segmentation, and object detection. CNNs are the foundational building blocks of all these vision-related tasks. Over the years, CNNs have evolved, becoming increasingly sophisticated in terms of structure, depth, and scalability~\cite{Krizhevsky2012, He2015, tan2020efficientnet}. 

Initially introduced for NLP, attention mechanisms have also proven effective in the CV field, enhancing the processing capabilities for complex image data~\cite{Vaswani2017}. A notable implementation in CV is through non-local operations, which employ a self-attention concept to capture long-term dependencies without the sequential information flow typical in multiple CNN layers, thereby without "bottom-up" feature extraction. Non-local operations achieve this by computing a weighted sum of all positions within the feature maps, thereby enhancing the vision architecture's ability to utilize global contextual information across the entire input field ~\cite{Wang_2018_CVPR}. Subsequently, attention mechanisms were more formally integrated into CV models. For instance, axial attention, which computes self-attention along a specified axis to evaluate similarities among elements within that axis, was an early adaptation for visual tasks~\cite{wang2020axialdeeplab, Ho2019}. Attention components can also be integrated into a typical CNN network to enhance feature extraction capabilities~\cite{park2018bam}.

A significant advancement in this field was the introduction of the ViT, which applies transformer architectures to image understanding tasks, offering a faster alternative to traditional region-based models while maintaining competitive performance~\cite{Dosovitskiy2020}. Despite the advantages of ViT and its variants, techniques focusing on local neighborhoods, such as CNN-based modules, are still incorporated into these models for their added value. In hybrid ViT, CNNs are used to extract visual tokens that serve as inputs for transformers, integrating the strengths of both architectures~\cite{wu2020visual, Dosovitskiy2020}. Additionally, CNNs can be embedded directly into the feed-forward networks within transformers to enhance their local processing capabilities~\cite{li2021localvit}. Moreover, the concept of a shifted window in CNNs has been adapted to transformer structures in the Swin Transformer~\cite{liu2021swin}, making it as a versatile backbone for CV tasks thanks to its ability to generate hierarchical feature maps as CNNs.

When it comes to solar forecasting, the CNN encoder was extensively explored. Sarkis \etal~\cite{Sarkis2024} leverage a model with CNN and LSTM layers using two public cameras in close proximity as a pair, demonstrating the forecasting potential of public cameras for the next 10 to 180 minutes. There are also similar explorations on all-sky images~\cite{Talha2019, zhang2018deep, paletta2020}. It has been transitioning to transformers in recent years. Jiri \etal~\cite{Jiri2022} utilized a GPT-2-based encoder applied to a sequence of frames for predicting GHI 15 minutes ahead. Likewise, Gao and Liu~\cite{Gao2022} employed a ViT to encode sky videos from the past hour (6 frames at 10-minute intervals) to predict GHI for the next 10 minutes up to 4 hours. These applications typically require a good camera to generate continuous frames representing time components. To develop a more lightweight, efficient and generalized model, we shifted to a different data modality: time series. This shift is alighed with a broader trend in CV field, as exemplified by the advancements in multimodal learning.

\textbf{Multimodel learning.} In multimodal learning, vision-and-language integration is a prevalent task. Data from different modalities are processed through modality-specific models to extract relevant features. Commonly, image encoders are CNN-backed~\cite{huang2020pixel}, while RNNs or transformers process sequential data, typically text. A robust ability to generate efficient visual embeddings is crucial, especially for tasks that demand object recognition and region of interest (RoI) extraction, such as visual question answering, visual captioning, or image-text retrieval~\cite{Kim2021, lu2019vilbert}. Later, it has been shown that simpler methods, such as using linear projections to extract visual features, can also achieve satisfying performance, as demonstrated in the work of Vision-and-Language Transformer (ViLT)~\cite{Kim2021}.

Cross-modal integration can occur at various stages. Early fusion might involve concatenating input data from different modalities at the entry level, while late fusion often utilizes heavy embedders to process different data modalities before the final integration. In previous work, lightweight modality interaction steps—such as simple concatenation, dot multiplication, or the use of shallow neural networks—were commonly employed, paired with heavy embedders to balance model complexity and performance~\cite{lee2018stacked, Radford2021}. However, recent research highlights the effectiveness of a deeper merge during modality interaction, typically employing transformer-based approaches~\cite{Kim2021, li2021align}.

Due to the unique characteristics of intraday solar forecasting tasks using public camera images and historical time series data, we developed SMT, an end-to-end predictor using a single-stream network. This method processes data into vectors via linear projection and feeds them into a transformer encoder at the early stage, inspired by the ViLT structure~\cite{Kim2021}. Our experiments suggest that two-stream networks using heavier visual embedders, such as CNNs or U-nets~\cite{ronneberger2015u}, are less effective for this specific task due to the relevant RoI, dataset size, and model size.
\section{Framework}
\begin{figure*}
    \centering
    \includegraphics[width=\linewidth]{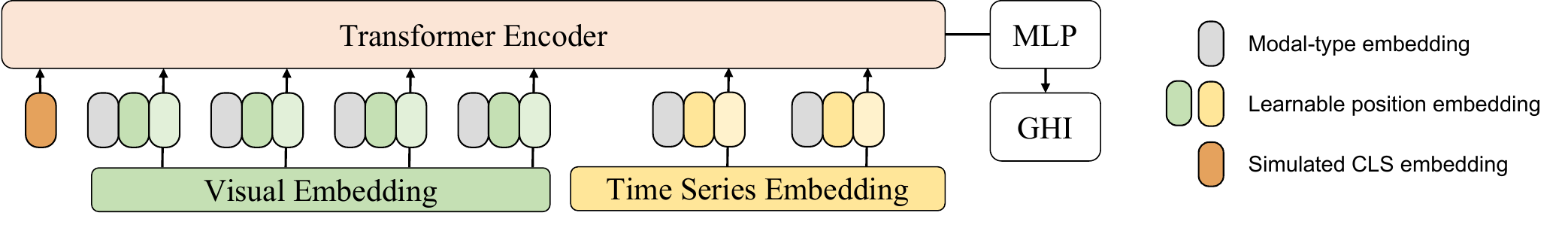}
    \caption{Model overview of Solar Multimodal Transformer. Illustration inspired by Kim \etal~\cite{Kim2021}. This end-to-end, single-stream model uses a basic transformer encoder to integrate data from multiple modalities. It linearly projects patches from images and time series data (historical GHI and optional meteorological data), before feeding them into the transformer for information fusion. The model employs a typical class token [CLS] to extract crucial information for the final prediction.}
    \label{fig:model}
\end{figure*}
\textbf{Problem formulation and evaluation metrics.} Solar forecasting can be formulated to predict the expected value of GHI at horizon $h$, denoted as $\hat{y}_{t+h}:=\mathbf{E}[y_{t+h}|x_t, y_{t-w+1:t},z_{t-w+1:t}]$. Here, $x_t$ represents the public image taken at time $t$, $y_{t-w+1:t}$ denotes a window of length $w$ of historical GHI time series, and $z_{t-w+1:t}$ are other optional meteorological vectors. All experiments in this paper target a short-term horizon of 2 hours. 

The accuracy of the prediction can be evaluated using three conventional metrics: root mean square error (RMSE), root relative squared error (RSE), and correlation (CORR). These metrics are defined as follows,
\begin{gather}
    RMSE = \sqrt{\frac{1}{n} \sum_{i=1}^n (y_i - \hat{y}_i)^2},\\
    RSE = \sqrt{\frac{\sum_{i=1}^n (y_i - \hat{y}_i)^2}{\sum_{i=1}^n (y_i - \overline{y})^2}},\\
    \text{CORR} = \frac{\sum_{i=1}^n (y_i - \overline{y})(\hat{y}_i - \overline{\hat{y}})}{\sqrt{\sum_{i=1}^n (y_i - \overline{y})^2 \sum_{i=1}^n (\hat{y}_i - \overline{\hat{y}})^2}},
\end{gather}
where $i$ belongs to the testing set $\Omega_{\text{test}}$ of size $n$, $y_i$ and $\hat{y}_i$ represent the observed and predicted values, while $\overline{y}$  and $\overline{\hat{y}}$ are the means of the observed and predicted values, respectively. The RMSE is used to represent the square root of the average squared differences between predicted and observed values, expressed in units of $W/m^2$, making it suitable for comparing the performance of models within the same testing set. The RSE, a scaled index of RMSE that accounts for the volatility of the predicted time series, provides a more reliable metric when comparing results across different testing sets. CORR measures the linear relationship between the observed and predicted series. Lower RMSE and RSE, and a higher CORR are expected.

\textbf{Architecture of SMT.} SMT adopts a structure similar to ViLT~\cite{Kim2021}, but with a modification: the text embedder is replaced by a time series data pillar. Its architecture is illustrated in \cref{fig:model} and described as following, 
\begin{subequations}
\begin{align}
    z_{img,0} &= [x_p^1\mathbf{E}_{m};...;x_p^N\mathbf{E}_{m}] + \mathbf{E}_{pos}^{img} + x^{type} \label{1_1}\\
    z_{ts,0} &= [ts^1\mathbf{E}_{l};...;ts^M\mathbf{E}_{l}] + \mathbf{E}_{pos}^{ts} + ts^{type} \label{1_2}\\
    z_0 &= [y_{prediction};z_{img,0};z_{ts,0}] \label{1_3}\\
    z'_{l} &= MAS(LN(z_{l-1})) + z_{l-1}, l = 1,..., L \label{1_4}\\
    z_{l} &= MLP(LN(z'_{l})) + z'_{l} , l = 1,..., L \label{1_5}\\
    \hat{y}_i &= z_{L}^{0}\mathbf{V} \label{1_6}
\end{align}
\end{subequations}

The entire network is composed of stacked blocks, each with multi-head self-attention (MSA) layers, a multilayer perceptron (MLP) head, and layer normalization (LN) embedded within them. The input image with channel number $C$, height $H$ and width $W$, $\mathbf{X} \in \mathbb{R}^{H \times W \times C}$, is initially partitioned into $N$ patches $x_p \in \mathbb{R}^{a \times b \times C}$, where $(a \times b)$ represents the patch size and $N = (H \times W) / (a \times b)$. These patches are then flattened and linearly projected using the visual embedding matrix, $\mathbf{E}_{m} \in \mathbb{R}^{C \times D}$, to produce visual vectors $v := x_p\mathbf{E}_{m} \in \mathbb{R}^{N \times D}$, where $D$ is the universal embedding size across all transformer blocks. These vectors $v$ are subsequently summed with a learnable positional embedding, $\mathbf{E}_{pos}^{img}$, for each patch, along with a fixed modal-type indicator for images, $x^{type}$.

Similarly, for the time series modality, if $M$ input time series $ts \in \mathbb{R}^{M \times S}$ are present (with $M=1$ if only use historical GHI time series), the embedding matrix, $\mathbf{E}_{l} \in \mathbb{R}^{S \times D}$, is used for linear projection. Here, $\mathbf{E}_{pos}^{ts}$ denotes the positional embedding for different vectors of the time series, and $ts^{type}$ serves as the modality type indicator with a fixed value.

The initial image feature representation $z_{img,0} \in \mathbb{R}^{N \times D}$ and the initial time series feature representation $z_{ts,0} \in \mathbb{R}^{M \times D}$ are concatenated with a learnable vector token $y_{prediction} \in \mathbb{R}^{1 \times D}$. This vector token is analogous to the class token [CLS] commonly used in NLP transformer settings, where the final prediction is derived. Once all inputs are merged into $z_0 \in \mathbb{R}^{(N+M+1)\times D}$, they are fed into the transformer block collectively with a residual connection, as shown in \cref{1_4,1_5}. By employing an MLP head with $\mathbf{V} \in \mathbb{R}^{D \times 1}$ to the first index of $z_{L}$, the regression output is extracted from what the simulated [CLS] token has learned.

\textbf{Hybrid SMT.} The linear projection of patches in the visual embedder can be replaced by other feature extractors, which are expected to introduce image-specific inductive biases beneficial for our regression task. We propose two variants: CNN + SMT, where the CNN latent space is extracted from a pretrained CNNLSTM model~\cite{Sarkis2024}, and U-net + SMT, where U-net feature maps are derived from a pretrained U-net model. Specifically, all skip connections in the U-net are eliminated to prevent direct information flow, forcing all data into the deepest latent space. This operation creates a ``dense" representation of the original image at the bottleneck, which is then fed into the transformer to integrate with other data modalities. Details of the implementation can be found in the supplementary.


\textbf{Forecasting with normalized GHI.} Clear sky models provide theoretical estimates of GHI values in the absence of clouds. We employ a simplified clear sky model, Haurwitz model, that solely considers the position of the Sun, i.e. the solar zenith angle~\cite{Haurwitz1945}, to generate clear sky GHI estimates. These estimates are used to normalize the GHI to a theoretical range between 0 and 1, as represented by:
\begin{equation}
  y^{\star}_{t} = \frac{y_t}{C^{\text{max clear sky}}_{day(t)}},
  \label{eq:scaled_ghi}
\end{equation}
where $C^{\text{max clear sky}}_{day(t)}$ denotes the daily maximal clear sky GHI value of the day for time $t$. This reference is preferred over $C^{\text{clear sky}}_{t}$ because it mitigates the risk of generating outliers during normalization, particularly if $C^{\text{clear sky}}_{t}$ is exceedingly small or zero. This normalization process helps to detrend the data over the long term, removing the influence of seasonal variations but rather on the degree of sky clearness, and simultaneously reflecting the relative time of day.

\section{Experiments}
\label{sec:experiments}
\subsection{Data} 
We use five datasets, detailed in Table \ref{tab:data}, each comprising public camera images paired with localized GHI data measured nearby. These datasets are collected from diverse geographical conditions, marked as urban open space (OO), urban streetscape (SS), valley, lake, and mountain. All public camera images are downloaded in real-time from Roundshot's website\footnote{\url{https://www.roundshot.com/en/home/livecam-references.html/102}} every 10 minutes---a company specializing in 360-degree panorama webcams. There are occasional missing values due to intermittent device malfunctions. Originally in high resolution, all panoramas are resized to the standard dimensions of $3 \times 224 \times 224$ for all experiments presented in this paper. Corresponding GHI data is collected from available weather stations operated by MeteoSwiss or nearby pyranometers. Further details about the dataset are available in the supplementary material.
\begin{table}[ht]
\centering

\begin{tabular}{p{1.5cm}p{1cm}p{1.5cm}p{0.9cm}p{1cm}}
\toprule
\textbf{Camera} & \textbf{\# Img} & \textbf{Range} & \textbf{Dis.(m)} & \textbf{Alt.(m)} \\ 
\midrule
Urban(OO) & 38676 & 2021.12.6-2024.1.30 & 125 & 405 \\ 
Urban(SS) & 31731 & 2023.07.05-2024.8.30 & 1180 & 420 \\ 
Valley & 57840 & 2022.07.28-2024.8.30 & 270 & 1447 \\ 
Lake & 36482 & 2023.04.10-2024.8.30 & 2000 & 440 \\ 
Mountain & 58389 & 2022.05.13-2024.8.30 & 300 & 2120 \\ 
\bottomrule
\end{tabular}

\caption{Summary of public camera datasets: This includes the landform where the camera is located, the number of image-GHI pairs collected, the date range of the collection, the distance between the camera and the GHI measurement location, and the altitude of each camera.}

\label{tab:data}
\end{table}
\subsection{Implementation details}
The presented framework is built using the Pytorch platform and the timm library~\cite{rw2019timm, Touvron2020}. We perform random search for hyperparameter optimization. Initialization and hyperparameters are set as follows: patch size $(a,b) = (16,16)$, the encoder layer $L = 3$, universal embedding size $D = 192$, number of heads $=6$, and batch size $=32$. Both the positional encodings and the simulated [CLS] token are one-dimensional and learnable, initialized with a normal distribution having a standard deviation of 0.02. We employ the AdamW optimizer, setting the learning rate at $5E-4$ with cosinus annealling at a ratio of 0.5 over 100 epochs, and a warm-up phase of 2 epochs starting from $5E-5$. Our objective is to minimize the mean squared error (MSE),  denoted as $L(y, \hat{y}) = \lVert y-\hat{y} \rVert ^2 $. Early stopping is applied after 20 epochs if no improvement is observed in the validation set to prevent overfitting. We train our models from scratch due to the unique nature of our task, and a detailed explanation can be found in the supplementary. 

\subsection{Evaluation of SMT}
\subsubsection{Comparison to state of the art}

\begin{table*}[ht]
\centering

\resizebox{\textwidth}{!}{
\begin{tabular}{llllllllllllllll}
\toprule
\multirow{2}{*}{\textbf{Model}} & \multicolumn{3}{c}{\textbf{Urban (open space)}} & \multicolumn{3}{c}{\textbf{Urban (streetscape)}} & \multicolumn{3}{c}{\textbf{Valley}} & \multicolumn{3}{c}{\textbf{Lake}} & \multicolumn{3}{c}{\textbf{Mountain}} \\
\cmidrule(lr){2-4} \cmidrule(lr){5-7} \cmidrule(lr){8-10} \cmidrule(lr){11-13} \cmidrule(lr){14-16}
& RMSE & RSE & CORR & RMSE & RSE & CORR & RMSE & RSE & CORR & RMSE & RSE & CORR & RMSE & RSE & CORR \\
\midrule
Persistent model~\cite{pedro2012} & 90.437 & 0.530 & 0.866  & 95.769 & 0.564 & 0.840 & 133.331 & 0.723 & 0.712 & 88.533 & 0.559 & 0.849 & 84.388 & 0.465 & 0.893 \\
CNNLSTM, 2cam~\cite{Sarkis2024} & 101.613 & 0.574 & 0.874 & - & - & - & - & - & - & - & - & - & - & - & - \\
CNNLSTM, 1cam~\cite{Sarkis2024} & 101.912 & 0.576 & 0.870 & 130.647 & 0.770 & 0.793 & 150.523 & 0.817 & 0.745 & 135.741 & 0.857 & 0.771 & 91.919 & 0.506 & 0.878 \\
LSTNet~\cite{Lai2017} & 76.010 & 0.445 & 0.899 & 113.075 & 0.666 & 0.790 & 107.670 & 0.584 & 0.815 & 106.002 & 0.669 & 0.823 & 86.393 & 0.476 & 0.880 \\
SMT & \textbf{65.434} & \textbf{0.383} & \textbf{0.927} & \textbf{86.070} & \textbf{0.507} & \textbf{0.871} & \textbf{79.996} & \textbf{0.434} & \textbf{0.901} & \textbf{71.935} & \textbf{0.454} & \textbf{0.892} & \textbf{74.253} & \textbf{0.409} & \textbf{0.914} \\
\bottomrule
\end{tabular}
}
\caption{Comparison to benchmark models across five datasets: Testing period ranges from September 27, 2023, to January 30, 2024, with a forecasting resolution of every 10 minutes and a prediction horizon of 2 hours. This testing period includes diverse weather conditions such as cloudy, partially-cloudy, and sunny days, which ensures a robust evaluation set. The model ``CNNLSTM, 2camera" is implemented exclusively on the ``urban open space" dataset, where a second nearby camera is available.}

\label{tab:eval}
\end{table*}
\textbf{Benchmarks.} To evaluate the performance of SMT, we adopted both some state-of-the-art deep learning models and ready-made industry predictions as benchmarks.
\begin{enumerate}
\item \textbf{Persistent model.} The smart persistence model, as proposed by Pedro \etal~\cite{pedro2012}, is straightforward yet accurate for short-horizon forecasting. It has proven superior to statistical models such as ARIMA applied to time series, or naive CNN approaches applied to images~\cite{Sarkis2024}. This physical model assumes that the degree of cloudiness persists over short periods, making it a reliable benchmark for our evaluations, as
\begin{equation}
  \hat{y}_{t+h} = C^{\text{clear sky}}_{t+h}*\frac{y_t}{C^{\text{clear sky}}_{t}}.
  \label{eq:smart}
\end{equation}

\item \textbf{CNNLSTM.} This is an image-based deep learning model from Sarkis \etal~\cite{Sarkis2024}, using a CNN + LSTM network to process a pair of current frames from adjacent cameras. We maintain the same setting for datasets where a second camera is available; for datasets with only one camera, we modify the network to process only a single frame for forecasting.

\item \textbf{LSTNet.} Proposed by Lai \etal~\cite{Lai2017}, LSTNet is a deep learning time series model that employs CNN and GRU components to capture both long- and short-term temporal patterns in solar irradiance. It has demonstrated significant improvement over some state-of-the-art baseline time series models, showing good performance on a benchmark dataset on solar energy\footnote{NREL, \url{https://www.nrel.gov/grid/solar-power-data.html}}.

\item \textbf{Solcast.} Solcast\footnote{\url{https://solcast.com/}} is a company specializing in solar irradiance forecasting, offering predictions ranging from a few minutes up to 14 days. They utilize geostationary satellite images and weather model data to provide global-scale irradiance forecasts. As Solcast’s methods are proprietary and not open-source, we directly compare our forecasts with their predictions to benchmark our model's performance. 
\end{enumerate}

\begin{figure}
  \centering
  \includegraphics[width=\linewidth]{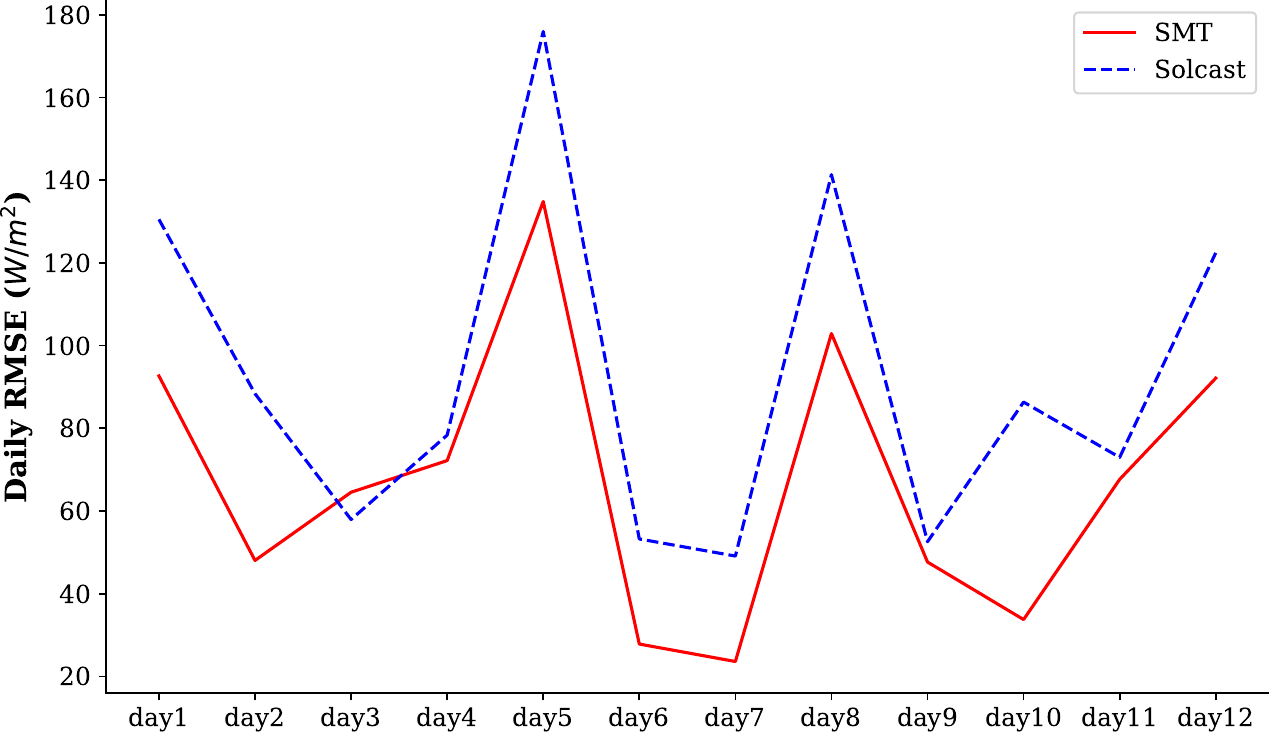}
    \caption{SMT \vs Solcast, daily RMSE}
    \label{fig:solcast}
\end{figure}

From the literature, evaluation metrics are typically reported for studies that forecast throughout the entire day. However, the GHI should theoretically be zero at night, and these nighttime zero-values, which constitute nearly half of the dataset, represent easier predictive tasks and can artificially inflate accuracy metrics. It is crucial to note that our datasets contain only daytime images. Thus, we applied the same datasets to these models for a fair comparison.

\Cref{tab:eval} summarizes the evaluation results of the SMT model. Across all the datasets, which vary in geographical conditions, weather patterns, and camera settings (such as focal length and depth of field), SMT consistently outperforms all other benchmark models. For further comparison with industry predictions, we collected real-time predictions from Solcast by providing the coordinates of the ``urban open space" camera to their API. Data was gathered every 10 minutes for a 2-hour prediction horizon over a period of 12 days, from March 28, 2023, to April 11, 2023, albeit with some gaps. The results are depicted in \cref{fig:solcast}, where the SMT exhibits a much lower daily RMSE throughout the entire 12-day testing period.\footnote{This testing period is excluded from the training samples during the training of SMT.} As a whole, the SMT achieves a RMSE of 76.943, outperforming Solcast’s RMSE 103.913 by 25.95\%, highlighting the effectiveness of our approach in practical applications. In addition, although the results reported here are for a prediction horizon of 2 hours, we list the performance of other horizons ranging from 10 minutes to 2 hours in the supplementary materials, where a shorter horizon correlates with better performance, as expected. 

\subsubsection{Ablation study}
\label{sec:ablation}
\textbf{GHI normalization analysis.} Ablation tests were conducted to isolate the impact of the GHI normalization step on the SMT. \Cref{tab:norm} shows that normalization generally helps with forecasting and has a larger effect on the image pillar by providing a sense of time to the forecaster.
\begin{table}[ht]
\centering

\begin{tabular}{lllll}
\toprule
\textbf{Model}  &\textbf{Norm}  &\textbf{RMSE} &\textbf{RSE} &\textbf{CORR} \\
\midrule
LSTNet~\cite{Lai2017} & - & 76.010 & 0.445 & 0.899 \\
LSTNet~\cite{Lai2017} & \checkmark & 72.921 & 0.427 & 0.906 \\
CNNLSTM~\cite{Sarkis2024} & - & 101.912 & 0.576 & 0.870\\
CNNLSTM~\cite{Sarkis2024} & \checkmark & 69.033 & 0.408 & 0.915\\
SMT & - & 78.281 & 0.458 & 0.889 \\
SMT & \checkmark & 65.434 & 0.383 & 0.927\\
\bottomrule
\end{tabular}
\caption{Impact of GHI normalization step. Tested from 28 Sep. 2023 to 29 Jan. 2024, on the ``OO" dataset.``Norm" denotes whether the GHI normalization step is applied.}
\label{tab:norm}
\end{table}

\begin{table}[ht]
\centering

\begin{tabular}{cccccccc}
\hline
\multirow{2}{*}{Idx}&\multicolumn{3}{c}{\textbf{Ablation}} & \multicolumn{3}{c}{\textbf{Performance}}\\
\cmidrule(lr){2-4} \cmidrule(lr){5-7}
 & pr & ts & img & RMSE & RSE & CORR \\ 
\midrule
\raisebox{.5pt}{\textcircled{\raisebox{-.9pt} {1}}} &linear & -   & 1 & 71.301 & 0.407 & 0.914 \\ 
\raisebox{.5pt}{\textcircled{\raisebox{-.9pt} {2}}} & linear & -   & 2 & 71.623 & 0.409 & 0.913 \\
\raisebox{.5pt}{\textcircled{\raisebox{-.9pt} 3}} & linear & -   & 3 & 73.482 & 0.420 & 0.908 \\
\raisebox{.5pt}{\textcircled{\raisebox{-.9pt} {4}}} & linear & 24h & -   & 79.698 & 0.467 & 0.886 \\
\midrule
\raisebox{.5pt}{\textcircled{\raisebox{-.9pt} {5}}} & linear & 30min & 1   & 71.624 & 0.409 & 0.913\\
\raisebox{.5pt}{\textcircled{\raisebox{-.9pt} {6}}} & \textbf{linear} &  \textbf{24h}& \textbf{1} & \textbf{65.434} & \textbf{0.383} & \textbf{0.927} \\ 
\midrule
\raisebox{.5pt}{\textcircled{\raisebox{-.9pt} {7}}} & CNN    & 24h & 1 & 66.681 & 0.390 & 0.921 \\ 
\raisebox{.5pt}{\textcircled{\raisebox{-.9pt} {8}}} & U-net  & 24h & 1 & 69.945 & 0.410 & 0.916 \\ 
\midrule
\raisebox{.5pt}{\textcircled{\raisebox{-.9pt} {9}}} & -  & ~\cite{Lai2017} & ~\cite{Sarkis2024} & 69.890 & 0.409 & 0.913\\ 
\bottomrule
\end{tabular}
\caption{Ablation study of the SMT, tested from 28 Sep. 2023 to 29 Jan. 2024, on the ``OO" dataset. ``pr" refers to the projection method of patches into vectors before being fed into the transformer; ``ts" indicates whether the time series embedder is used; ``img" denotes whether the visual embedder is used. The GHI normalization step is applied in all cases. SMT is highlighted in bold.}
\label{tab:ablation}
\end{table}
\textbf{Model components.} We perform ablations on different components of the SMT, as shown in Table \ref{tab:ablation}. Model \raisebox{.5pt}{\textcircled{\raisebox{-.9pt} {1}}}-\raisebox{.5pt}{\textcircled{\raisebox{-.9pt} {3}}} use only the image component of the SMT (ViT) with a stack of frames to capture temporal information. For example, model {\textcircled{\raisebox{-.9pt} {3}}} processes frames $X_{t-20min},X_{t-10min},X_{t} $ as inputs to predict $y_{t+120min}$.  The modality type token is kept to differentiate patches from different frames. Model {\textcircled{\raisebox{-.9pt} {4}}} uses only the time series component of SMT. Model {\textcircled{\raisebox{-.9pt} {5}}}, a SMT with historical GHI for the past 30 minutes, serves as a counterpart to model \raisebox{.5pt}{\textcircled{\raisebox{-.9pt} {3}}}, aligning our experiments with literature that uses multi-frame inputs (or videos). The rationale for excluding the time series data was to assess whether temporal dynamics could be sufficiently captured solely through the stack of frames. However, the results showed no improvement with an increased number of continuous frames. This is not surprising. In fact, before the adoption of two-stream networks~\cite{simonyan2014two}, it has been challenging to learn spatio-temporal features from video frames as demonstrated by Karpathy \etal~\cite{karpathy2014large}. Moreover, unlike in universal action recognition tasks in CV field, the relationship between inputs and outputs in solar prediction tasks is not direct. Specifically, the movement of clouds, as captured by continuous frames, does not visually correlate with solar irradiance, necessitating additional analytical steps such as direction identification and spatial construction, which are challenging to learn from a limited dataset.

Moreover, model \raisebox{.5pt}{\textcircled{\raisebox{-.9pt} {5}}} shows that the added value of the past 30-minute GHI is not readily apparent, whereas model \raisebox{.5pt}{\textcircled{\raisebox{-.9pt} {6}}}, the SMT using 24-hour historical GHI outperformed the alternatives. This underscores the importance of the time series data's window length. Given that our camera doesn't operate at night, a more appropriate future comparison test would involve frames spanning the past 24 hours, which would provide a direct comparison with model {\textcircled{\raisebox{-.9pt} {6}}}.

Hybrid SMTs like model \raisebox{.5pt}{\textcircled{\raisebox{-.9pt} {7}}} and \raisebox{.5pt}{\textcircled{\raisebox{-.9pt} {8}}} tend to converge more quickly due to the bottom-up features input into the transformers. However, they are also prone to overfitting. Introducing a dropout rate before the final linear layer of the transformer or within the attention blocks has not significantly mitigated this issue, a limitation possibly attributed to the size of our dataset.

Model \raisebox{.5pt}{\textcircled{\raisebox{-.9pt} {9}}} combine LSTNet~\cite{Lai2017} and CNNLSTM (1 camera)~\cite{Sarkis2024} through a simple concatenation step before making the final prediction. Despite the combination of these two strong models on individual data modalities, their performance is still less effective than that of the SMT, demonstrating the effectiveness of the transformer component for a deep integration of spatial and temporal information from diverse data modalities in this task. Although the ViT has been observed to inadequately model fine details in images, leading to a loss of feature richness due to its simple tokenization of image patches~\cite{yuan2021tokenstotoken}, this limitation is not a concern for solar forecasting using public camera images. In the case of public cameras, the majority of the captured scene, such as buildings and vegetation, remains static, containing largely irrelevant information. Unlike tasks such as semantic segmentation, which require recognition of local structures like edges or lines, our focus is primarily on dynamic elements such as optical flow, cloud movements, and the position of the Sun, etc. Furthermore, the images from public cameras are panoramic, stretched out into a long rectangular format. The different areas of the image are related but not directly connected; for example, pixels representing regions at $0^{\circ}$ North and $180^{\circ}$ South are in the same line of sight but appear disconnected in the image. The transformer architecture enables the establishment of dependencies right from the outset between pixels that are distantly located yet highly interdependent, in contrast to the ``bottom-up" feature aggregation after layers of CNNs.

\subsubsection{Patch shape and attention analysis}
\label{sec:patch}
\begin{figure*}
    \centering
    \includegraphics[width=\linewidth]{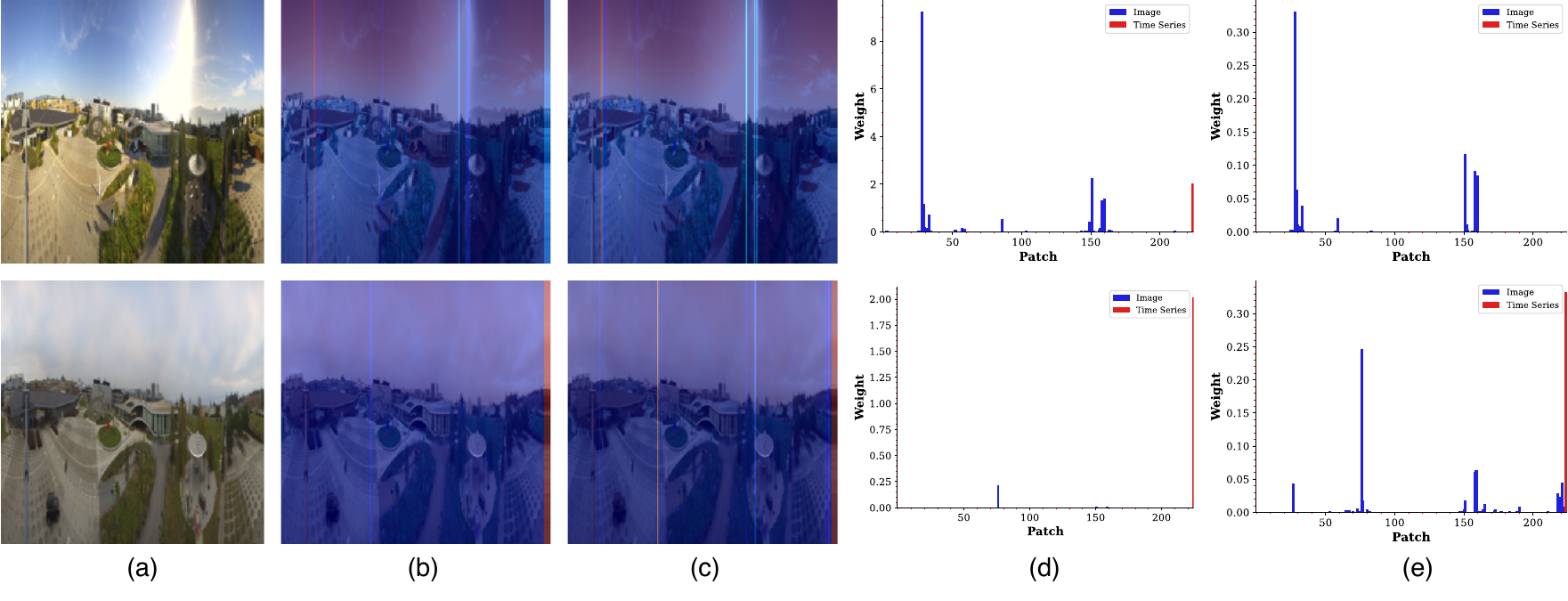}
    \caption{Attention analysis using column patches for SMT: (a) Unprocessed images. (b) Patch-specific visualizations using attention weighted rollout. (c) Patch-specific visualizations from the last attention block. (d) Bar plots of weighted attention, quantifying the influence of different patches throughout the model’s layers. (e) Bar plots of attention from the last layer, illustrating the final focus before a forecast is made. The accentuated patches in panels (b, c) indicate areas of higher importance for the final prediction. To incorporate the contribution of time series along with column patches in these visualizations, panels (b, c) include an added column of pixels at the end. Similarly, the time series component is visualized as the last vector in panels (d, e), marked in red, to highlight its relative importance.}
    \label{fig:attnsum}
\end{figure*}
To investigate how patch shapes influence the performance of SMT, we experimented with modifying the patch shapes to row or column pixels, in addition to the standard square patches of $16 \times 16$. Column patches that carry directional information, are proved to be most effective, see Table \ref{tab:patch_comparison}. 

\begin{table}[ht]
\centering
\begin{tabular}{llll}
\toprule
\textbf{Patch}  &\textbf{RMSE} &\textbf{RSE} &\textbf{CORR} \\
\midrule
Square & 65.434 &0.383 &0.927  \\
Row & 66.112 &0.387 &0.923   \\
Column & 63.756 &0.373 &0.928   \\
\bottomrule
\end{tabular}
\caption{The influence of different patch shapes for SMT. Tested from 28 Sep. 2023 to 29 Jan. 2024, on the ``OO" dataset.}
\label{tab:patch_comparison}
\end{table}
The attention maps help us understand the critical information relied upon for final predictions. To visualize how transformers interpret these patches, we not only show the attention from the last layer, but also employ a weighted attention rollout methodology~\cite{Chefer_2021_CVPR}, which involves multiplying the attention map of each layer by its gradients. In our experiments with column patches as shown in \cref{fig:attnsum}, we note distinct behaviors under varying weather conditions. Specifically, in sunny conditions, especially when the Sun is visible, the regressor actively tracks the Sun's location and its diametrically opposite direction, even though these areas appear irrelevant in the panorama. This observation suggests that the model efficiently utilizes directional information from the imagery to predict solar irradiance. Conversely, under mostly cloudy conditions where direct sunlight is obscured, historical GHI data becomes significantly more influential for forecasting. This shift indicates the model's adaptive reliance on available data types based on environmental cues. Comparing the bar plots of weighted and last layer attention, we notice that the fusion of different data modalities primarily occurs in the initial, shallower blocks of the transformer. This early-stage integration allows for the effective combination of immediate image data and historical time series, while the final layer of the transformer predominantly focuses on refining the learning directly from the images. More examples are provided in the supplementary materials.

In our analysis of attention visualization using row patches, we observe that the regressor utilizes not only segments from the sky but also those from the ground. This finding aligns with results as Sarkis \etal~\cite{Sarkis2024}, which showed that excluding ground pixels---rich in localized information such as reflections, shadows, and sunlight---adversely affects the model's performance. 

The insights gained from the attention maps substantiate the effectiveness of using a single transformer encoder for our specific task. Typically, in multimodal structures such as vision-and-language pre-training tasks, performance heavily depends on visual feature extraction, which is usually achieved through convolutional structures and region supervision. However, for our task, the visual information is less complex and does not demand high expressive power. Consequently, a single transformer encoder is sufficient to capture this information efficiently. This simpler set-up allows the model to focus more effectively on the fusion of different data modalities, streamlining the process while maintaining robust performance.
\section{Conclusion}
We introduced a novel solution for localized intraday solar forecasting, which consists of three parts. First, we explored the potential of utilizing single-frame imagery from public cameras for solar forecasting. Second, we applied a normalization technique tailored for solar irradiance time-series data. Last, we presented the SMT, a lightweight multimodal model that integrates these elements and maximizes the forecasting ability. Benchmarking against the solar forecasting company Solcast, SMT demonstrated a 25.95\% improvement in RMSE over a 12-day testing period in April 2023. This model eliminates the need for paired images or video data from cameras ~\cite{Sarkis2024,Talha2019,Gao2022}, while delivering even better performance. The pretrained SMT can be easily fine-tuned to adapt to any public cameras with varying focal lengths or depths, offering extensive potential for broader applications.

\section*{Acknowledgment}
We sincerely appreciate EPFL's Solutions for Sustainability Initiative (S4S) grant. We also extend our special thanks to Solcast for supplying the essential data.

{\small
\bibliographystyle{ieee_fullname}
\bibliography{egbib}

\begin{thebibliography}{10}\itemsep=-1pt

\bibitem{Bi2023}
Kaifeng Bi, Lingxi Xie, Hengheng Zhang, Xin Chen, Xiaotao Gu, and Qi Tian.
\newblock Accurate medium-range global weather forecasting with 3d neural
  networks.
\newblock {\em Nature}, 619:533--538, 7 2023.

\bibitem{Chefer_2021_CVPR}
Hila Chefer, Shir Gur, and Lior Wolf.
\newblock Transformer interpretability beyond attention visualization.
\newblock In {\em Proceedings of the IEEE/CVF Conference on Computer Vision and
  Pattern Recognition (CVPR)}, pages 782--791, June 2021.

\bibitem{chung2014empirical}
Junyoung Chung, Caglar Gulcehre, KyungHyun Cho, and Yoshua Bengio.
\newblock Empirical evaluation of gated recurrent neural networks on sequence
  modeling.
\newblock {\em arXiv preprint arXiv:1412.3555}, 2014.

\bibitem{de2015error}
Maria~Grazia De~Giorgi, Paolo~Maria Congedo, Maria Malvoni, and Domenico
  Laforgia.
\newblock Error analysis of hybrid photovoltaic power forecasting models: A
  case study of mediterranean climate.
\newblock {\em Energy conversion and management}, 100:117--130, 2015.

\bibitem{Dong2021}
Bing Dong, Reisa Widjaja, Wenbo Wu, and Zhi Zhou.
\newblock Review of onsite temperature and solar forecasting models to enable
  better building design and operations.
\newblock {\em Building Simulation}, 14:885--907, 2021.

\bibitem{Dosovitskiy2020}
Alexey Dosovitskiy, Lucas Beyer, Alexander Kolesnikov, Dirk Weissenborn,
  Xiaohua Zhai, Thomas Unterthiner, Mostafa Dehghani, Matthias Minderer, Georg
  Heigold, Sylvain Gelly, Jakob Uszkoreit, and Neil Houlsby.
\newblock An image is worth 16x16 words: Transformers for image recognition at
  scale, 10 2020.

\bibitem{Gao2022}
Huiyu Gao and Miaomiao Liu.
\newblock Short-term solar irradiance prediction from sky images with a clear
  sky model.
\newblock In {\em 2022 IEEE/CVF Winter Conference on Applications of Computer
  Vision (WACV)}, pages 3074--3082, 2022.

\bibitem{Guermoui2018}
Mawloud Guermoui, Farid Melgani, and Céline Danilo.
\newblock Multi-step ahead forecasting of daily global and direct solar
  radiation: A review and case study of ghardaia region.
\newblock {\em Journal of Cleaner Production}, 201:716--734, 2018.

\bibitem{Haurwitz1945}
Bernhard Haurwitz.
\newblock Insolation in relation to cloudiness and cloud density.
\newblock {\em Journal of The Atmospheric Sciences - J ATMOS SCI}, 2:154--166,
  01 1945.

\bibitem{He2015}
Kaiming He, Xiangyu Zhang, Shaoqing Ren, and Jian Sun.
\newblock Deep residual learning for image recognition, 12 2015.

\bibitem{Hewamalage2021}
Hansika Hewamalage, Christoph Bergmeir, and Kasun Bandara.
\newblock Recurrent neural networks for time series forecasting: Current status
  and future directions.
\newblock {\em International Journal of Forecasting}, 37:388--427, 1 2021.

\bibitem{Ho2019}
Jonathan Ho, Nal Kalchbrenner, Dirk Weissenborn, and Tim Salimans.
\newblock Axial attention in multidimensional transformers, 12 2019.

\bibitem{hochreiter1997long}
Sepp Hochreiter and J{\"u}rgen Schmidhuber.
\newblock Long short-term memory.
\newblock {\em Neural computation}, 9(8):1735--1780, 1997.

\bibitem{huang2020pixel}
Zhicheng Huang, Zhaoyang Zeng, Bei Liu, Dongmei Fu, and Jianlong Fu.
\newblock Pixel-bert: Aligning image pixels with text by deep multi-modal
  transformers.
\newblock {\em arXiv preprint arXiv:2004.00849}, 2020.

\bibitem{ineichen2006comparison}
Pierre Ineichen.
\newblock Comparison of eight clear sky broadband models against 16 independent
  data banks.
\newblock {\em Solar Energy}, 80(4):468--478, 2006.

\bibitem{karpathy2014large}
Andrej Karpathy, George Toderici, Sanketh Shetty, Thomas Leung, Rahul
  Sukthankar, and Li Fei-Fei.
\newblock Large-scale video classification with convolutional neural networks.
\newblock In {\em Proceedings of the IEEE conference on Computer Vision and
  Pattern Recognition}, pages 1725--1732, 2014.

\bibitem{Kim2021}
Wonjae Kim, Bokyung Son, and Ildoo Kim.
\newblock Vilt: Vision-and-language transformer without convolution or region
  supervision, 2 2021.

\bibitem{Krizhevsky2012}
Alex Krizhevsky, Ilya Sutskever, and Geoffrey Hinton.
\newblock Imagenet classification with deep convolutional neural networks.
\newblock {\em Neural Information Processing Systems}, 25, 01 2012.

\bibitem{Lai2017}
Guokun Lai, Wei-Cheng Chang, Yiming Yang, and Hanxiao Liu.
\newblock Modeling long- and short-term temporal patterns with deep neural
  networks, 3 2017.

\bibitem{lecun1995convolutional}
Yann LeCun, Yoshua Bengio, et~al.
\newblock Convolutional networks for images, speech, and time series.
\newblock {\em The handbook of brain theory and neural networks},
  3361(10):1995, 1995.

\bibitem{lee2018stacked}
Kuang-Huei Lee, Xi Chen, Gang Hua, Houdong Hu, and Xiaodong He.
\newblock Stacked cross attention for image-text matching.
\newblock In {\em Proceedings of the European conference on computer vision
  (ECCV)}, pages 201--216, 2018.

\bibitem{li2021align}
Junnan Li, Ramprasaath Selvaraju, Akhilesh Gotmare, Shafiq Joty, Caiming Xiong,
  and Steven Chu~Hong Hoi.
\newblock Align before fuse: Vision and language representation learning with
  momentum distillation.
\newblock {\em Advances in neural information processing systems},
  34:9694--9705, 2021.

\bibitem{li2021localvit}
Yawei Li, Kai Zhang, Jiezhang Cao, Radu Timofte, and Luc~Van Gool.
\newblock Localvit: Bringing locality to vision transformers, 2021.

\bibitem{liu2021swin}
Ze Liu, Yutong Lin, Yue Cao, Han Hu, Yixuan Wei, Zheng Zhang, Stephen Lin, and
  Baining Guo.
\newblock Swin transformer: Hierarchical vision transformer using shifted
  windows, 2021.

\bibitem{lu2019vilbert}
Jiasen Lu, Dhruv Batra, Devi Parikh, and Stefan Lee.
\newblock Vilbert: Pretraining task-agnostic visiolinguistic representations
  for vision-and-language tasks.
\newblock {\em Advances in neural information processing systems}, 32, 2019.

\bibitem{paletta2020}
Quentin Paletta and Joan Lasenby.
\newblock Convolutional neural networks applied to sky images for short-term
  solar irradiance forecasting.
\newblock {\em arXiv preprint arXiv:2005.11246}, 2020.

\bibitem{paletta2023advances}
Quentin Paletta, Guillermo Terr{\'e}n-Serrano, Yuhao Nie, Binghui Li, Jacob
  Bieker, Wenqi Zhang, Laurent Dubus, Soumyabrata Dev, and Cong Feng.
\newblock Advances in solar forecasting: Computer vision with deep learning.
\newblock {\em Advances in Applied Energy}, page 100150, 2023.

\bibitem{park2018bam}
Jongchan Park, Sanghyun Woo, Joon-Young Lee, and In~So Kweon.
\newblock Bam: Bottleneck attention module.
\newblock {\em arXiv preprint arXiv:1807.06514}, 2018.

\bibitem{pedro2012}
Hugo~TC Pedro and Carlos~FM Coimbra.
\newblock Assessment of forecasting techniques for solar power production with
  no exogenous inputs.
\newblock {\em Solar Energy}, 86(7):2017--2028, 2012.

\bibitem{Jiri2022}
Jiří Pospíchal, Martin Kubovčík, and Iveta~Dirgová Luptáková.
\newblock Solar irradiance forecasting with transformer model.
\newblock {\em Applied Sciences (Switzerland)}, 12, 9 2022.

\bibitem{Radford2021}
Alec Radford, Jong~Wook Kim, Chris Hallacy, Aditya Ramesh, Gabriel Goh,
  Sandhini Agarwal, Girish Sastry, Amanda Askell, Pamela Mishkin, Jack Clark,
  Gretchen Krueger, and Ilya Sutskever.
\newblock Learning transferable visual models from natural language
  supervision, 2 2021.

\bibitem{ronneberger2015u}
Olaf Ronneberger, Philipp Fischer, and Thomas Brox.
\newblock U-net: Convolutional networks for biomedical image segmentation.
\newblock In {\em Medical image computing and computer-assisted
  intervention--MICCAI 2015: 18th international conference, Munich, Germany,
  October 5-9, 2015, proceedings, part III 18}, pages 234--241. Springer, 2015.

\bibitem{Sarkis2024}
Roy Sarkis, Ilker Oguz, Demetri Psaltis, Mario Paolone, Christophe Moser, and
  Luisa Lambertini.
\newblock Intraday solar irradiance forecasting using public cameras.
\newblock {\em Solar Energy}, 275:112600, 6 2024.

\bibitem{Talha2019}
Talha~A. Siddiqui, Samarth Bharadwaj, and Shivkumar Kalyanaraman.
\newblock A deep learning approach to solar-irradiance forecasting in
  sky-videos, 2019.

\bibitem{simonyan2014two}
Karen Simonyan and Andrew Zisserman.
\newblock Two-stream convolutional networks for action recognition in videos.
\newblock {\em Advances in neural information processing systems}, 27, 2014.

\bibitem{tan2020efficientnet}
Mingxing Tan and Quoc~V. Le.
\newblock Efficientnet: Rethinking model scaling for convolutional neural
  networks, 2020.

\bibitem{Tascikaraoglu2018}
Akin Tascikaraoglu.
\newblock Evaluation of spatio-temporal forecasting methods in various smart
  city applications.
\newblock {\em Renewable and Sustainable Energy Reviews}, 82:424--435, 2018.

\bibitem{Touvron2020}
Hugo Touvron, Matthieu Cord, Matthijs Douze, Francisco Massa, Alexandre
  Sablayrolles, and Hervé Jégou.
\newblock Training data-efficient image transformers \& distillation through
  attention, 12 2020.

\bibitem{Vaswani2017}
Ashish Vaswani, Noam Shazeer, Niki Parmar, Jakob Uszkoreit, Llion Jones, Aidan
  Gomez, Lukasz Kaiser, and Illia Polosukhin.
\newblock Attention is all you need, 06 2017.

\bibitem{wang2020axialdeeplab}
Huiyu Wang, Yukun Zhu, Bradley Green, Hartwig Adam, Alan Yuille, and
  Liang-Chieh Chen.
\newblock Axial-deeplab: Stand-alone axial-attention for panoptic segmentation,
  2020.

\bibitem{Wang_2018_CVPR}
Xiaolong Wang, Ross Girshick, Abhinav Gupta, and Kaiming He.
\newblock Non-local neural networks.
\newblock In {\em Proceedings of the IEEE Conference on Computer Vision and
  Pattern Recognition (CVPR)}, June 2018.

\bibitem{rw2019timm}
Ross Wightman.
\newblock Pytorch image models.
\newblock \url{https://github.com/rwightman/pytorch-image-models}, 2019.

\bibitem{wu2020visual}
Bichen Wu, Chenfeng Xu, Xiaoliang Dai, Alvin Wan, Peizhao Zhang, Zhicheng Yan,
  Masayoshi Tomizuka, Joseph Gonzalez, Kurt Keutzer, and Peter Vajda.
\newblock Visual transformers: Token-based image representation and processing
  for computer vision, 2020.

\bibitem{yuan2021tokenstotoken}
Li Yuan, Yunpeng Chen, Tao Wang, Weihao Yu, Yujun Shi, Zihang Jiang, Francis~EH
  Tay, Jiashi Feng, and Shuicheng Yan.
\newblock Tokens-to-token vit: Training vision transformers from scratch on
  imagenet, 2021.

\bibitem{zhang2015baseline}
Jie Zhang, Bri-Mathias Hodge, Siyuan Lu, Hendrik~F Hamann, Brad Lehman, Joseph
  Simmons, Edwin Campos, Venkat Banunarayanan, Jon Black, and John Tedesco.
\newblock Baseline and target values for regional and point pv power forecasts:
  Toward improved solar forecasting.
\newblock {\em Solar Energy}, 122:804--819, 2015.

\bibitem{zhang2018deep}
Jinsong Zhang, Rodrigo Verschae, Shohei Nobuhara, and Jean-Fran{\c{c}}ois
  Lalonde.
\newblock Deep photovoltaic nowcasting.
\newblock {\em Solar Energy}, 176:267--276, 2018.

\end{thebibliography}
}

\end{document}